\documentclass[letterpaper]{article} 

\usepackage{rotating}
\usepackage{threeparttable}
\usepackage{adjustbox}
\usepackage{subfigure}
\usepackage{caption}
\usepackage{graphicx}
\usepackage{float} 
\usepackage{subcaption}

\usepackage{booktabs}
\usepackage{amssymb}
\usepackage{lscape}
\usepackage{multirow}
\usepackage{xcolor}
\usepackage{amsmath} 
\usepackage{textcomp}
\usepackage{pifont} 

\definecolor{customgreen}{HTML}{d5eded}
\definecolor{customred}{HTML}{f9e1d0}

\usepackage{aaai2026}
\usepackage{times}  
\usepackage{helvet}  
\usepackage{courier}  
\usepackage[hyphens]{url}  
\usepackage{graphicx} 
\usepackage{natbib}  
\usepackage{caption} 
\usepackage{algorithm}
\usepackage{algorithmic}
\usepackage{url}

\usepackage{newfloat}
\usepackage{listings}
\DeclareCaptionStyle{ruled}{labelfont=normalfont,labelsep=colon,strut=off} 
\floatstyle{ruled}
\newfloat{listing}{tb}{lst}{}
\floatname{listing}{Listing}
\title{Empowering Drug Discovery through Intelligent, Multitask Learning and OOD Generalization}

\author {
    Kun Li\textsuperscript{\rm 1}\equalcontrib,
    WuZhennan\textsuperscript{\rm 1}\equalcontrib,
    Yida Xiong\textsuperscript{\rm 1}\equalcontrib,
    Hongzhi Zhang\textsuperscript{\rm 1}\equalcontrib,
    Hu Longtao\textsuperscript{\rm 1}\equalcontrib,
    Zhonglie Liu\textsuperscript{\rm 1}\equalcontrib,
    Junqi Zeng\textsuperscript{\rm 1}\equalcontrib,
    Wenjie Wu\textsuperscript{\rm 1}\equalcontrib,
    Junqi Zeng\textsuperscript{\rm 2}\equalcontrib,
    Zhonglie Liu\textsuperscript{\rm 1}\equalcontrib,
    Longtao Hu\textsuperscript{\rm 1}\equalcontrib,
    Mukun Chen\textsuperscript{\rm 1}\equalcontrib,
    Jiameng Chen\textsuperscript{\rm 1},
    Wenbin Hu\textsuperscript{\rm 1}\thanks{Corresponding author}
}

\affiliations {
    \textsuperscript{\rm 1}School of Computer Science, Wuhan University, Wuhan 430037, Hubei, China\\
    \textsuperscript{\rm 2}Key Laboratory of Aerospace Information Security and Trusted Computing,\\
    Ministry of Education, School of Cyber Science and Engineering, Wuhan University, Wuhan 430037, Hubei, China\\
    likun98@whu.edu.cn, yidaxiong@whu.edu.cn, zhanghongzhi@whu.edu.cn, wjwu111@whu.edu.cn, zengjunqi@whu.edu.cn,\\
    zlliucs20210820@whu.edu.cn, hlt\_2003@whu.edu.cn, wuzhennan@whu.edu.cn, cmk0910@whu.edu.cn,\\
    jiameng.chen@whu.edu.cn, hwb@whu.edu.cn
}

\usepackage{bibentry}

\begin{document}

\maketitle

\begin{abstract}

Drug discovery is of great social significance in safeguarding human health, prolonging life, and addressing the challenges of major diseases. In recent years, artificial intelligence has demonstrated remarkable advantages in key tasks across bioinformatics and pharmacology, owing to its efficient data processing and data representation capabilities. However, most existing computational platforms cover only a subset of core tasks, leading to fragmented workflows and low efficiency. In addition, they often lack algorithmic innovation and show poor generalization to out-of-distribution (OOD) data, which greatly hinders the progress of drug discovery. To address these limitations, we propose \underline{B}ai\underline{s}heng\underline{l}ai (\textbf{BSL}), a deep learning-enhanced, open-access platform designed for virtual drug discovery. BSL integrates seven core tasks within a unified and modular framework, incorporating advanced technologies such as generative models and graph neural networks. In addition to achieving state-of-the-art (SOTA) performance on multiple benchmark datasets, the platform emphasizes evaluation mechanisms that focus on generalization to OOD molecular structures. Comparative experiments with existing platforms and baseline methods demonstrate that BSL provides a comprehensive, scalable, and effective solution for virtual drug discovery, offering both algorithmic innovation and high-precision prediction for real-world pharmaceutical research. In addition, BSL demonstrated its practical utility by discovering novel modulators of the GluN1/GluN3A NMDA receptor, successfully identifying three compounds with clear bioactivity in \textit{in-vitro} electrophysiological assays. These results highlight BSL as a promising and comprehensive platform for accelerating biomedical research and drug discovery. The platform is accessible at https://www.baishenglai.net.

\end{abstract}


\section{Introduction}



With the rapid advancement of biomedicine and computational technology, the field of drug discovery is undergoing profound transformations \cite{mishra2022stroke, mock2023ai}. Traditional drug development processes are typically lengthy and costly, often taking 10 to 15 years and requiring investments in the billions of dollars \cite{congreve2014structure, catacutan2024machine, lavecchia2024advancing}. However, such substantial expenditures do not guarantee success, primarily due to the high failure rate in clinical trials, where many candidate drugs fail to demonstrate the expected efficacy or safety. These challenges exacerbate the difficulties associated with drug development. To address these issues, computer-aided drug design (CADD) \cite{cadd} has emerged, with virtual screening technologies playing a critical role in drug selection and optimization.


\begin{figure*}[ht!]
    \centering
    \includegraphics[width=0.995\linewidth]{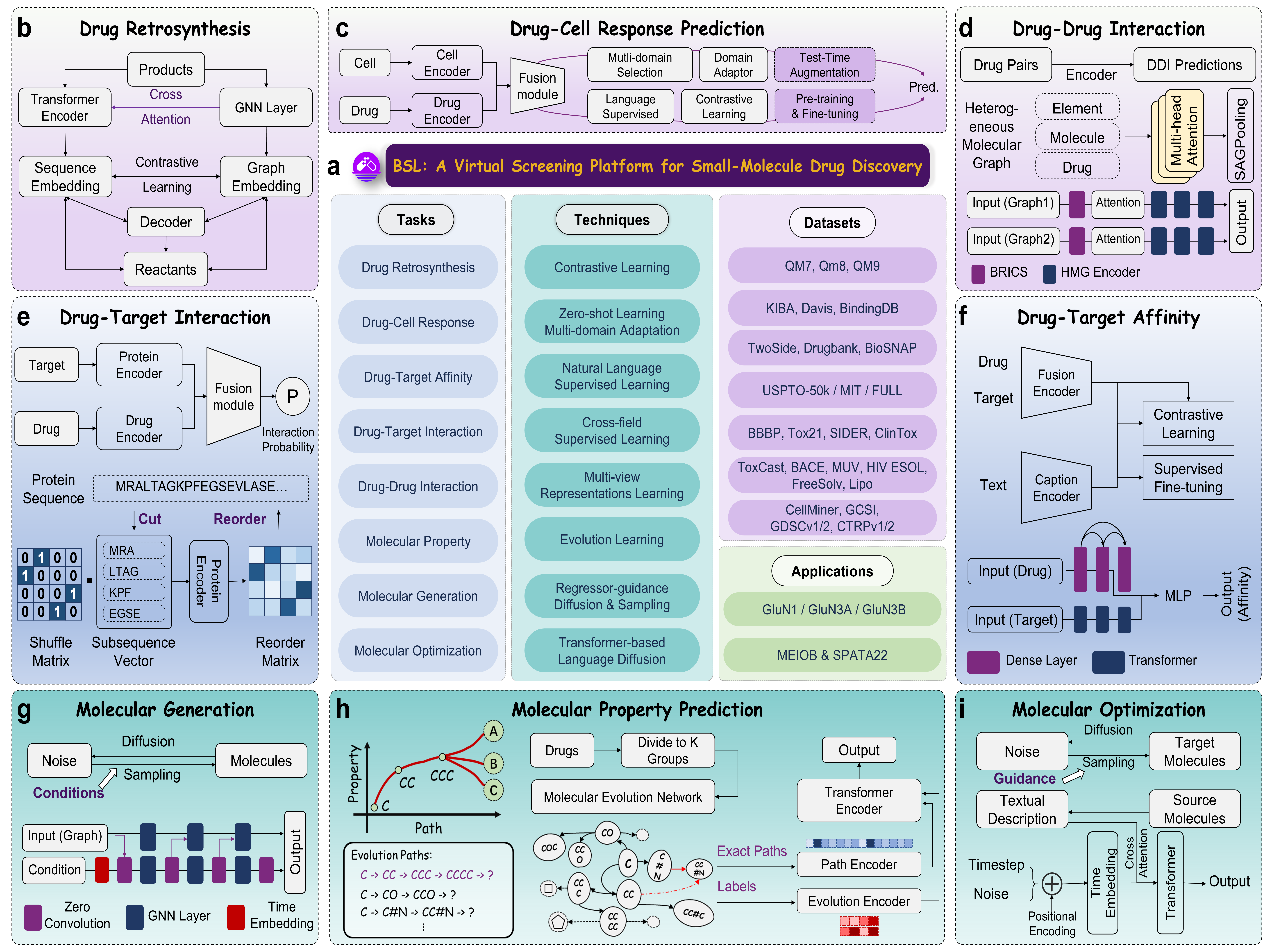}
    \caption{The core design of the BSL platform, including its supported tasks, underlying technologies, benchmark datasets, and downstream applications.  \textbf{b}–\textbf{i} present the technical workflows tailored to each specific task.}
    \label{fig:main}
\end{figure*}

In recent years, many drug-related virtual screening platforms have emerged as computational technology has advanced. These platforms can be roughly categorized into several types. The first type is simulation-based platforms, which rely on molecular docking and dynamics to model ligand–target interactions. For example, Schrödinger offers a comprehensive suite of tools for molecular modeling, binding affinity prediction, and structure-based drug design. Similarly, OpenVS \cite{zhou2024artificial} enables local simulation of drug–protein interactions, providing flexibility for deployment in specific research environments. The second type comprises deep learning-based platforms \cite{shen2024drugflow}, which leverage large-scale datasets and AI models to predict molecular properties and biological activity.  For instance, iDrug\footnote{https://drug.ai.tencent.com/cn} integrates a range of prediction methods to meet broader screening needs, making them more versatile and adaptable for drug discovery. These platforms significantly accelerate early-stage drug development by enabling rapid identification of promising candidates, reducing experimental costs, and improving the efficiency and accuracy of virtual screening and prediction workflows.

Although these platforms demonstrate significant advantages in drug discovery, they still face numerous limitations in practical applications. For example, DrugFlow \cite{shen2024drugflow} has limited support for out-of-distribution (OOD) data, and its screening tasks are restricted to a predefined set of 4,317 molecular libraries and specific target pockets, making it difficult to adapt to open-ended drug design scenarios. Furthermore, many platforms rely on fixed datasets for model training, neglecting the model's generalization ability for OOD data. When faced with structurally novel or clinically unseen compounds, model performance can significantly decline, thereby reducing prediction accuracy and reliability. More seriously, current platforms also have obvious shortcomings in task completeness. Although some platforms can complete basic property prediction tasks, molecular generation, and target-based drug screening, they lack supporting algorithms and data for tasks such as drug-cell response prediction, drug retrosynthesis, and drug-drug interaction prediction, making it difficult to provide systematic solutions. Additionally, the functional fragmentation and inconsistent data structure standards across multiple platforms often require researchers to manually convert data formats and integrate results, significantly increasing technical barriers and collaboration costs. Some platforms also have limitations on uploading custom data, charging for use, and having closed-source code, further raising the barriers to access. These limitations have had a tangible impact on modern drug discovery. They slow down the progression from target identification to lead compound selection, leading to inefficient use of resources, and hindering the practical deployment of AI-assisted design. Critically, the lack of platform reliability and flexibility contributes to the high failure rate in early-stage drug development and limits the industry's transition toward more intelligent, efficient paradigms. This underscores the urgent need for an end-to-end drug design platform with strong integration, generalization, and task extensibility to meet the demands of real-world biomedical challenges.

\begin{table*}[t!]
    \centering
    \renewcommand\arraystretch{0.6}
    \resizebox{\textwidth}{!}{
    \begin{tabular}{lccccccc|ccccccccc}
         \toprule
        \multirow[c]{2}{*}{Platform}& \multicolumn{7}{c|}{Tasks} & \multicolumn{5}{c}{Features} \\
        \cline{2-13}
         & \raisebox{-0.9ex}{MG} & \raisebox{-0.9ex}{MO} & \raisebox{-0.9ex}{MPP} & \raisebox{-0.9ex}{DTI} & \raisebox{-0.9ex}{DDI} & \raisebox{-0.9ex}{DRP} & \raisebox{-0.9ex}{Retro} & \raisebox{-0.9ex}{Public} & \raisebox{-0.9ex}{Free} & \raisebox{-0.9ex}{OOD} & \raisebox{-0.9ex}{AI+} & \raisebox{-0.9ex}{Innov.}  \\
        \midrule
        DrugFlow \cite{shen2024drugflow}        & \colorbox{customgreen}{\ding{51}} & \colorbox{customgreen}{\ding{51}} & \colorbox{customgreen}{\ding{51}} & \colorbox{customgreen}{\ding{51}} & \colorbox{customred}{\ding{55}} & \colorbox{customred}{\ding{55}} & \colorbox{customred}{\ding{55}} & \colorbox{customgreen}{\ding{51}} & \colorbox{customgreen}{\ding{51}} & \colorbox{customgreen}{\ding{51}} & 4 & 4  \\
        MolProphet \cite{keda2024molprophet}   & \colorbox{customgreen}{\ding{51}} & \colorbox{customgreen}{\ding{51}} & \colorbox{customred}{\ding{55}} & \colorbox{customred}{\ding{55}} & \colorbox{customred}{\ding{55}} & \colorbox{customred}{\ding{55}} & \colorbox{customred}{\ding{55}} & \colorbox{customgreen}{\ding{51}} & \colorbox{customgreen}{\ding{51}} & \colorbox{customgreen}{\ding{51}} & 2 & 2  \\
        AIDDISON \cite{rusinko2023aiddison}        & \colorbox{customgreen}{\ding{51}} & \colorbox{customred}{\ding{55}} & \colorbox{customgreen}{\ding{51}} & \colorbox{customred}{\ding{55}} & \colorbox{customgreen}{\ding{51}} & \colorbox{customred}{\ding{55}} & \colorbox{customgreen}{\ding{51}} & \colorbox{customred}{\ding{55}} & \colorbox{customred}{\ding{55}} & \colorbox{customgreen}{\ding{51}} & 4 & 3  \\
        Chemistry42 \cite{ivanenkov2023chemistry42}     & \colorbox{customgreen}{\ding{51}} & \colorbox{customgreen}{\ding{51}} & \colorbox{customgreen}{\ding{51}} & \colorbox{customred}{\ding{55}} & \colorbox{customred}{\ding{55}} & \colorbox{customred}{\ding{55}} & \colorbox{customred}{\ding{55}} & \colorbox{customred}{\ding{55}} & -  & \colorbox{customgreen}{\ding{51}} & 3 & 3 \\
        PanGu \cite{lin2022pangu}      & \colorbox{customgreen}{\ding{51}} & \colorbox{customgreen}{\ding{51}} & \colorbox{customgreen}{\ding{51}} & \colorbox{customgreen}{\ding{51}} & \colorbox{customred}{\ding{55}} & \colorbox{customred}{\ding{55}} & \colorbox{customgreen}{\ding{51}} & \colorbox{customred}{\ding{55}} & \colorbox{customred}{\ding{55}} & \colorbox{customred}{\ding{55}} & 5 & 5 \\
        DockThor-VS \cite{guedes2024dockthor}     & \colorbox{customred}{\ding{55}} & \colorbox{customred}{\ding{55}} & \colorbox{customgreen}{\ding{51}} & \colorbox{customgreen}{\ding{51}} & \colorbox{customred}{\ding{55}} & \colorbox{customred}{\ding{55}} & \colorbox{customred}{\ding{55}} & \colorbox{customgreen}{\ding{51}} & \colorbox{customred}{\ding{55}} & \colorbox{customgreen}{\ding{51}} & 4 & 4 \\
        EXSCALATE \cite{gadioli2022extreme}       & \colorbox{customgreen}{\ding{51}} & \colorbox{customred}{\ding{55}} & \colorbox{customred}{\ding{55}} & \colorbox{customgreen}{\ding{51}} & \colorbox{customred}{\ding{55}} & \colorbox{customred}{\ding{55}} & \colorbox{customred}{\ding{55}} & \colorbox{customred}{\ding{55}} & \colorbox{customred}{\ding{55}} & \colorbox{customgreen}{\ding{51}} & 4 & 4 \\
        CarbonAI \cite{wu2022carbonai}        & \colorbox{customgreen}{\ding{51}} & \colorbox{customgreen}{\ding{51}} & \colorbox{customgreen}{\ding{51}} & \colorbox{customgreen}{\ding{51}} & \colorbox{customred}{\ding{55}} & \colorbox{customred}{\ding{55}} & \colorbox{customred}{\ding{55}} & \colorbox{customred}{\ding{55}} & \colorbox{customred}{\ding{55}} & \colorbox{customgreen}{\ding{51}} & 6 & 6 \\
        CanDo \cite{fine2019computational}           & \colorbox{customred}{\ding{55}} & \colorbox{customgreen}{\ding{51}} & \colorbox{customgreen}{\ding{51}} & \colorbox{customgreen}{\ding{51}} & \colorbox{customred}{\ding{55}} & \colorbox{customred}{\ding{55}} & \colorbox{customred}{\ding{55}} & \colorbox{customgreen}{\ding{51}} & \colorbox{customgreen}{\ding{51}} & \colorbox{customgreen}{\ding{51}} & 3 & 3 \\
        iDrug \textsuperscript{1}           & \colorbox{customgreen}{\ding{51}} & \colorbox{customred}{\ding{55}} & \colorbox{customgreen}{\ding{51}} & \colorbox{customred}{\ding{55}} & \colorbox{customred}{\ding{55}} & \colorbox{customred}{\ding{55}} & \colorbox{customgreen}{\ding{51}} & \colorbox{customgreen}{\ding{51}} & \colorbox{customgreen}{\ding{51}} & \colorbox{customgreen}{\ding{51}} & 4 & 4 \\
        MOE \textsuperscript{2}             & \colorbox{customgreen}{\ding{51}} & \colorbox{customred}{\ding{55}} & \colorbox{customgreen}{\ding{51}} & \colorbox{customgreen}{\ding{51}} & \colorbox{customred}{\ding{55}} & \colorbox{customred}{\ding{55}} & \colorbox{customred}{\ding{55}} & \colorbox{customred}{\ding{55}} & \colorbox{customred}{\ding{55}} & \colorbox{customgreen}{\ding{51}} & 2 & 3 \\
        Cresset \textsuperscript{3}         & \colorbox{customgreen}{\ding{51}} & \colorbox{customgreen}{\ding{51}} & \colorbox{customred}{\ding{55}} & \colorbox{customgreen}{\ding{51}} & \colorbox{customred}{\ding{55}} & \colorbox{customred}{\ding{55}} & \colorbox{customred}{\ding{55}} & \colorbox{customgreen}{\ding{51}} & \colorbox{customred}{\ding{55}} & \colorbox{customgreen}{\ding{51}}  & 3 & 3\\
        BIOVIA \textsuperscript{4}         & \colorbox{customgreen}{\ding{51}} & \colorbox{customgreen}{\ding{51}} & \colorbox{customred}{\ding{55}} & \colorbox{customgreen}{\ding{51}} & \colorbox{customred}{\ding{55}} & \colorbox{customred}{\ding{55}} & \colorbox{customred}{\ding{55}} & \colorbox{customred}{\ding{55}} & \colorbox{customred}{\ding{55}} & \colorbox{customgreen}{\ding{51}} & 3 & 3 \\
        Eyesopen \textsuperscript{5}       & \colorbox{customgreen}{\ding{51}} & \colorbox{customgreen}{\ding{51}} & \colorbox{customred}{\ding{55}} & \colorbox{customred}{\ding{55}} & \colorbox{customred}{\ding{55}} & \colorbox{customred}{\ding{55}} & \colorbox{customred}{\ding{55}} & \colorbox{customred}{\ding{55}} & \colorbox{customred}{\ding{55}} & \colorbox{customgreen}{\ding{51}}  & 2 & 2\\
        Schrodinger \textsuperscript{6}    & \colorbox{customgreen}{\ding{51}} & \colorbox{customgreen}{\ding{51}} & \colorbox{customred}{\ding{55}} & \colorbox{customred}{\ding{55}} & \colorbox{customred}{\ding{55}} & \colorbox{customred}{\ding{55}} & \colorbox{customred}{\ding{55}} & \colorbox{customred}{\ding{55}} & \colorbox{customred}{\ding{55}} & \colorbox{customgreen}{\ding{51}}  & 2 & 2\\ \hline
        BSL (Ours) & \colorbox{customgreen}{\ding{51}} & \colorbox{customgreen}{\ding{51}} & \colorbox{customgreen}{\ding{51}} & \colorbox{customgreen}{\ding{51}} & \colorbox{customgreen}{\ding{51}} & \colorbox{customgreen}{\ding{51}} & \colorbox{customgreen}{\ding{51}} & \colorbox{customgreen}{\ding{51}} & \colorbox{customgreen}{\ding{51}} & \colorbox{customgreen}{\ding{51}} & \textbf{7} & \textbf{7} \\
        \bottomrule
    \end{tabular}}
    \parbox{\textwidth}{
        \small
        \textsuperscript{1}\url{https://drug.ai.tencent.com/cn.} \textsuperscript{2}\url{https://cloudscientific.com/product-drug-design-28.} \textsuperscript{3}\url{https://www.cresset-group.com/software.} \textsuperscript{4}\url{https://www.3ds.com/products/biovia/discovery-studio.} \textsuperscript{5}\url{https://www.eyesopen.com/orion/small-molecule-discovery-suite.} \textsuperscript{6}\url{https://www.schrodinger.com/life-science.}
    }
    \caption{Comparison of drug discovery platforms. \textbf{Public} indicates that the platform is currently available without the need for application or intervention from the service provider. \textbf{AI+} indicates the proportion of AI technology in drug generation, with higher scores indicating a higher level of AI technology integration. \textbf{Innov.} indicates the originality of the algorithms, with higher scores indicating higher originality of the platform's model algorithms.}
    \label{tab:comparison}
\end{table*}


Therefore, we developed the \underline{B}ai\underline{s}heng\underline{l}ai (\textbf{BSL}), a comprehensive virtual screening platform for drug discovery. As shown in Fig. \ref{fig:main}, BSL targets seven core tasks, including molecular condition generation and optimization, drug target affinity prediction, drug-cell response prediction, drug-drug interaction prediction, property prediction, and synthesis pathway prediction, equipped with 12 different deep learning enhancement methods. The platform extensively incorporates advanced techniques such as zero-shot learning, domain adaptation, diffusion models, graph neural networks, and contrastive learning, establishing a strong technical foundation. BSL effectively tackles practical challenges such as poor OOD generalization and difficulties in unifying multi-task data. It is freely accessible to the public and supports customized data analysis and prediction. We conducted fair and comprehensive benchmarking against recent representative methods across the seven tasks, and results show that BSL achieves SOTA performance on all tasks. In addition, BSL demonstrated its practical utility by discovering novel modulators of the GluN1/GluN3A NMDA receptor, successfully identifying three compounds with clear bioactivity in \textit{in vitro} electrophysiological assays. These results highlight BSL as a promising and comprehensive platform for accelerating biomedical research and drug discovery.



\section{Comparison with Other Web Platforms}

To position the BSL within the current drug discovery platform, we conducted a comparative analysis with representative web platforms along two dimensions: \textbf{task coverage} and \textbf{platform features}, as summarized in Table~\ref{tab:comparison}.

\paragraph{Task coverage.}
Platforms such as DrugFlow \cite{shen2024drugflow}, MolProphet \cite{keda2024molprophet}, and CarbonAI \cite{wu2022carbonai} are strong in specific subdomains. DrugFlow integrates molecular generation, docking, ADMET (absorption, distribution, metabolism, excretion and toxicity) prediction, and virtual screening, and supports lead identification and optimization pipelines. MolProphet provides structure-based and ligand-based virtual screening, AI‑based molecule generation, scaffold hopping, fragment/R‑group optimization, and docking assistance. CarbonAI employs a non‑docking, deep learning framework using GNN and transformer models. However, none of these platforms covers all seven key tasks. These omissions results in fragmented workflows that cannot handle full-cycle drug discovery.

\paragraph{Platform features.}
Most existing platforms employ AI-based solutions;  however, the majority are neither fully open-access nor freely available.  Even among accessible platforms such as iDrug \cite{congreve2014structure} and DrugFlow \cite{shen2024drugflow}, there are substantial limitations in supporting custom data uploads and analyses, along with the absence of a robust framework for generalizing to OOD data.  Industrial-grade platforms such as Schrödinger \cite{tang2022discoveryXDE}, BIOVIA, and Cresset rely on high-precision simulation-based methods and demonstrate strong predictive accuracy.  Nevertheless, these platforms are closed-source commercial systems, offering limited user accessibility and customization capabilities.  Their computational efficiency is often insufficient for large-scale virtual screening and the diverse range of downstream drug discovery tasks encountered in practical applications.

In summary, while existing web platforms offer specialized capacities in early stages, none match BSL in its combination of full-task coverage, deep AI integration, OOD generalization, and open accessibility. This comprehensive, modular design positions BSL as a more scalable, reliable, and practical choice for end-to-end virtual drug discovery.

\section{Overview of the Primary Functions}

\subsection{Molecular Generation Task}





Molecular generation is a crucial task in the field of drug discovery. The primary goal of this task is to design new compounds with potential for development into effective drugs. 
Existing condition-based molecular generation methods always face limitations in ensuring the effectiveness of the sampling space \cite{MG3,MG4}, leading to the generation of many ineffective molecules. To address the challenges faced by traditional methods, as illustrated in Fig.~\ref{fig:main}g, we incorporate RMCD into the BSL,  an innovative molecular generation method combining the score estimation of a diffusion model with the gradient from a regression controller model based on the target cell lines and IC50 scores. Meanwhile, we introduce a novel method for collaborative drug design named DDI-Diff \cite{huDDIRMCD}. DDI-Diff leverages known drug-drug interaction (DDI) information as conditional input to guide the diffusion process, enabling the generation of new candidate molecules with potential synergistic effects.
Since a molecule can be represented as a graph $G = (X, A)$, the sampling process of RMCD and DDI-Diff is described as follows: 
\begin{equation}
    \begin{aligned}
        & \left\{
        \begin{aligned}
            & \text{d}X_t = [\textbf{f}_{1,t}(X_t) - g^2_{1,t}\nabla_{X_t}\log{p_t}(X_t, A_t)]\text{d}\overline{t} + g_{1,t}\text{d}\overline{\textbf{w}}_1 \\
            & \text{d}A_t = [\textbf{f}_{2,t}(A_t) - g^2_{2,t}\nabla_{A_t}\log{p_t}(X_t, A_t)]\text{d}\overline{t} + g_{2,t}\text{d}\overline{\textbf{w}}_2, \\
        \end{aligned}
        \right.
    \end{aligned}
\end{equation}
where $\textbf{f}_{1,t}$ and $\textbf{f}_{2,t}$ are linear drift coefficients, $g_{1,t}$ and $g_{2,t}$ are scalar diffusion coefficients, and $\overline{\textbf{w}}_1$ and $\overline{\textbf{w}}_2$ are reverse-time standard Wiener processes.

\subsection{Molecular Optimization Task}

The task of conditional molecular optimization is critical for controllably designing and fine-tuning therapeutic compounds. In the BSL, we integrate novel methods focusing on conditional and multi-property molecular optimization. FMOP is a fragment-masked molecular optimization method based on phenotypic drug discovery \cite{li2024fragment}, which utilizes a regression-free diffusion model to conditionally optimize masked regions of molecules without training. It effectively generates new molecules with similar scaffolds. 
TransDLM \cite{xiong2024text} uses IUPAC \cite{iupac} names as semantic representations and encodes property requirements into text, reducing error propagation during the diffusion process, as shown in Fig.~\ref{fig:main}i. Its sampling procedure can be described as follows:
\begin{equation}
    q(x_{t} \mid x_{t-1}) = \mathcal{N}(x_{t}; \sqrt{1 - \beta_{t}} x_{t-1}, \beta_{t} I ),
\end{equation}
where $x_t$ denotes the embedding state of molecules at time $t$, $\beta_{t}$ is a hyperparameter representing the amount of noise added at diffusion step $t$.

\subsection{Molecular Property Prediction Task}

Molecular property prediction (MPP) estimates key chemical and biological attributes that determine a drug’s potential \cite{wieder2020compact} chemical properties. However, they often lack mechanisms to effectively leverage inter-property correlations or integrate external pharmacological knowledge. To tackle this complex task, BSL incorporates two novel methods. MEvoN \cite{mevon} is a molecular evolution network that creates unified molecular representations by combining a molecule's structural information with its evolutionary context. As shown in Fig.~\ref{fig:main}h, it achieves this by organizing molecules hierarchically, identifying evolutionary links based on similarity metrics, and then traversing these links to capture structural transitions and weighting them by known properties, all while encoding the molecule's intrinsic graph embedding. The evolutionary relationships are established by:
\begin{equation}
    \begin{aligned}
        & \left\{
        \begin{aligned}
            & \text{Pair}^1 = \{(m_p, m_q) | \mathcal{S}_{\text{edit/fp}}(m_p, m_q) \geq \theta_1\} \\
            & \text{Pair}^2 = \{(m'_p, m'_q) | \mathcal{S}_{\text{wl}}(m_p, m_q) \geq \theta_2\}, \\
        \end{aligned}
        \right.
    \end{aligned}
\end{equation}
where $m$ denotes any molecule, $\mathcal{S}_{\text{edit/fp}}$ is a function calculating fingerprint-based molecular similarity, $\mathcal{S}_{\text{wl}}$ calculates graph-based similarity using Weisfeiler-Lehman graph kernel, and $(m'_p, m'_q) \in \text{Pair}^1_{\text{max}}$ and $\text{Pair}^2$ is the final result of revolutionary relationships.
KCHML advances molecular property prediction by encoding molecular graphs into heterogeneous structures across three views—molecular, elemental, and pharmacological—integrating external knowledge for richer representations.

\begin{table*}[ht!]
\centering
\renewcommand\arraystretch{0.98}
\setlength{\tabcolsep}{5pt}
\resizebox{\textwidth}{!}{
\begin{tabular}{llccccc}
\toprule
Datasets & Metrics & \multicolumn{5}{c}{\colorbox{gray!20}{\textcolor{black}{Molecule Generation Task}}} \\[2pt]
\cmidrule(l){3-7}
& & CDGS $_{\text{\cite{huang2023conditional}}}$ & GruM-2D $_{\text{\cite{jo2023graph}}}$ & MOOD $_{\text{\cite{lee2023exploring}}}$ & RMCD (BSL) & \\
QM9 + & FCD $\downarrow$ & 77.0 / 61.1 / 53.1 & 85.3 / 60.7 / 52.7 & 80.2  / 57.7 / 48.7 & \textbf{77.0 / 56.0 / 47.8} & \\
GDSCv2 & MMD $\downarrow$ & .340 / .142 / .110 & .337 / .138 / .106 & .347 / .195 / .144 & \textbf{.313 / .142 / .101} & \\
\bottomrule 
\toprule
Datasets & Metrics & \multicolumn{5}{c}{\colorbox{gray!20}{\textcolor{black}{Molecular Optimization Task}}} \\[2pt]
\cmidrule(l){3-7}
& & Prompt-MolOpt $_{\text{\cite{wu2024leveraging}}}$ & HN-GFN $_{\text{\cite{zhu2023sample}}}$ & FFLOM $_{\text{\cite{jin2023fflom}}}$ & FMOP (BSL) & \\
QM9 +  & Success $\uparrow$ & 91.68\% & 92.70\% & 88.83\% & \textbf{95.43\%} & \\
GDSCv2 & Improv. $\uparrow$ & 5.70\% & 3.30\% & 6.3\% & \textbf{7.50\%} & \\
\cmidrule(l){3-7}
& & MolSearch $_{\text{\cite{sun2022molsearch}}}$ & FRATTVAE $_{\text{\cite{inukai2024tree}}}$ & DyMol $_{\text{\cite{shin2024dymol}}}$ & TransDLM (BSL) & \\
\multirow{1}{*}{ChEMBL} & FCD $\downarrow$ & 1.355 & 17.972 & 5.155 & \textbf{0.109} & \\
\bottomrule 
\toprule
Datasets & Metrics & \multicolumn{5}{c}{\colorbox{gray!20}{\textcolor{black}{Drug Target Interaction/Affinity Prediction Task}}} \\[2pt]
\cmidrule(l){3-7}
& & FOTF-CPI $_{\text{\cite{yin2024fotf}}}$ & HiGraphDTI $_{\text{\cite{liu2024higraphdti}}}$ & MGNDTI $_{\text{\cite{peng2024mgndti}}}$ & SiamDTI (BSL) & \\
BindingDB & \multirow{2}{*}{AUROC $\uparrow$} & 0.506 $\pm$ 0.030 & 0.540 $\pm$ 0.030 & 0.524 $\pm$ 0.032 & \textbf{0.554 $\pm$ 0.016} & \\
BioSNAP & & 0.590 $\pm$ 0.030 & 0.629 $\pm$ 0.030 & 0.601 $\pm$ 0.012 & \textbf{0.636 $\pm$ 0.020} & \\
\cmidrule(l){3-7}
& & RF $_{\text{\cite{ballester2010machine}}}$ & MSGNN-DTA $_{\text{\cite{wang2023msgnn}}}$ & HGRL-DTA $_{\text{\cite{chu2022hierarchical}}}$ & CLG-DTA (BSL) &\\
\multirow{2}{*}{KIBA} & PCC $\uparrow$ & 0.150 & 0.116 & 0.083 & \textbf{0.280} & \\
& MSE $\downarrow$ & 0.962 & 0.907 & 1.024 & \textbf{0.900} & \\
\bottomrule 
\toprule
Datasets & Metrics & \multicolumn{5}{c}{\colorbox{gray!20}{\textcolor{black}{Molecular Property Prediction Task}}} \\[2pt]
\cmidrule(l){3-7}
& & KCL $_{\text{\cite{fang2022molecular}}}$ & GROVER $_{\text{\cite{rong2020self}}}$ (iDrug) & KCHML (BSL) & MEvoN (BSL) \\
QM7 & \multirow{2}{*}{MAE $\downarrow$} & 59.9 & 90 & 56.1 & \textbf{45.9} \\
QM8 &  & 0.013 & 0.018 & \textbf{0.0121} & - \\
\cmidrule(l){3-7}
& & ComENet $_{\text{\cite{NEURIPS2022_0418973e}}}$ & Equiformer $_{\text{\cite{equiformer}}}$ & ViSNet $_{\text{\cite{visnet}}}$ & MEvoN (BSL) \\
QM9$_{\varepsilon_{\mathrm{HOMO}}}$ & \multirow{4}{*}{MAE $\downarrow$} & 0.0924 & 0.0263 & 0.0322 & \textbf{0.0223} \\
QM9$_{\varepsilon_{\mathrm{LUMO}}}$ &  & 0.0638 & 0.0236 & 0.0258 & \textbf{0.0233} \\
QM9$_{\Delta\varepsilon}$ &  & 0.1232 & 0.0492 & 0.0515 & \textbf{0.0227} \\
QM9$_{\mathrm{ZPVE}}$ &  & 0.0068 & 0.0019 & 0.0014 & \textbf{0.0005} \\
\bottomrule 
\toprule
Datasets & Metrics & \multicolumn{5}{c}{\colorbox{gray!20}{\textcolor{black}{Drug-Drug Interaction Prediction Task}}} \\[2pt]
\cmidrule(l){3-7}
& & GMPNN $_{\text{\cite{nyamabo2022drug}}}$ & DGNN-DDI $_{\text{\cite{ma2023dual}}}$& DSN-DDI $_{\text{\cite{li2023dsn}}}$ & KCHML (BSL) & \\
TwoSide$^{\textit{a}}$ & \multirow{2}{*}{AUC $\uparrow$} 
    & $77.69 \pm 0.26$ 
    & $77.25 \pm 0.23$ 
    & $77.25 \pm 0.23$ 
    & \textbf{81.87 $\pm$ 0.54} & \\
TwoSide$^{\textit{b}}$ & 
    & $80.91 \pm 0.80$ 
    & $81.96 \pm 0.26$ 
    & $81.59 \pm 0.66$ 
    & \textbf{83.75 $\pm$ 0.89} & \\
\cmidrule(l){3-7}
& & CDGS $_{\text{\cite{huang2023conditional}}}$ & MOOD $_{\text{\cite{lee2023exploring}}}$ & DruM-2D & DDI-diff (BSL) & \\
DrugBank, QM9 & ACC $\uparrow$ & 63.5 & 72.6 & 62.1 & \textbf{84.5} & \\
\bottomrule 
\toprule
Datasets & Metrics & \multicolumn{5}{c}{\colorbox{gray!20}{\textcolor{black}{Drug Response Prediction Task$^{\textit{c}}$}}} \\[2pt]
\cmidrule(l){3-7}
& & $\text{MSDA}_{\text{GraTransDRP}}$ & $\text{MSDA}_{\text{TransEDRP}}$ & $\text{CLDR}_{\text{GraTransDRP}}$  & $\text{CLDR}_{\text{TransEDRP}}$ \\
\multirow{2}{*}{GDSCv2} & PCC $\uparrow$ & 0.5103 (5.3\%) & 0.5316 (+5.1\%) & \textbf{0.5288 (+11.61\%)} & 0.5149 (+1.77\%) \\
& MSE $\downarrow$ & 0.0039 (+20.6\%) & 0.0044 (15.2\%) & 0.0039 (+17.02\%) & \textbf{0.0038 (+26.23\%)} \\
\bottomrule 
\toprule
Datasets & Metrics & \multicolumn{5}{c}{\colorbox{gray!20}{\textcolor{black}{Drug Retrosynthesis Task}}} \\[2pt]
\cmidrule(l){3-7}
& & GET-LT1 $_{\text{\cite{mao2021molecular}}}$ & SCROP $_{\text{\cite{zheng2019predicting}}}$ & RetroXpert $_{\text{\cite{yan2020retroxpert}}}$ & CFC-Retro (BSL) \\
\multirow{3}{*}{USPTO-50K} & Top-1 $\uparrow$ & 59.1 & 59.0 & 62.1 & \textbf{65.9} \\
& Top-3 $\uparrow$ & 73.4 & 74.8 & 75.8 & \textbf{80.4} \\
& Top-5 $\uparrow$ & 76.4 & 78.1 & 78.5 & \textbf{82.4} \\
\bottomrule
\end{tabular}
}
\parbox{\textwidth}{
    \scriptsize
    $^{\textit{a}}$: Both molecules unseen in training data. $^{\textit{b}}$: Only one molecule present in training. \\
    $^{\textit{c}}$: MSDA and CLDR are enhancement plug-ins introduced by BSL to improve the performance of baseline methods such as GraTransDRP $_{\text{\cite{chu2023graphtsdrp}}}$ and TransEDRP $_{\text{\cite{li2022transedrp}}}$. \\
}
\caption{Performance comparison across all tasks. The results demonstrate how BSL methods perform across different tasks, in comparison with the current SOTA methods. \textbf{Bold} indicates the best result.}
\label{tab:comparison2}
\end{table*}

\subsection{Drug Target Interaction/Affinity Prediction Task}

Drug–target interaction (DTI) and drug–target affinity (DTA) prediction tasks are typically formulated as classification and regression problems, respectively. However, existing methods are generally limited to in-distribution predictions and often fail to consider OOD generalization, particularly when encountering unseen drugs or targets. In the BSL, we integrate two advanced methods to respectively tackle DTI and DTA tasks, efficiently resolving the OOD issue. Siam-DTI utilizes a double-channel network structure for cross-field supervised learning, as shown in Fig.~\ref{fig:main}e. To capture both local and global protein information, a cross-field information fusion strategy is employed. Separately, as shown in Fig.~\ref{fig:main}f, CLG-DTA \cite{li2025contrastive} introduces a common-sense numerical knowledge graph (CN-KG), which leverages a linear structure to accurately and intuitively represent numerical values and their interrelationships. The CN-KG aims to minimize the difference between the embedding vectors of the entity set $E$ and the relationship set $L$:
\begin{equation}
    \mathcal{L}_{\text{KGE}} = \sum_{(h,l,t) \in \mathbb{S}} [ \gamma + \textbf{d}(h+l, t) - \textbf{d}(t+l, h) ]_+
\end{equation}
where $\textbf{d}$ denotes the similarity metric, $[x]_+$ denotes the positive part of $x$, and $\gamma > 0$ is the margin hyperparameter. The set $\mathbb{S}$ contains triplets $(h,l,t)$, where $h,t \in E$, and $l \in L$. For the similarity metric $\textbf{d}$, either the $L_1$ norm or the $L_2$ norm can be used.

\begin{figure*}[ht!]
    \centering
    \includegraphics[width=0.95\linewidth]{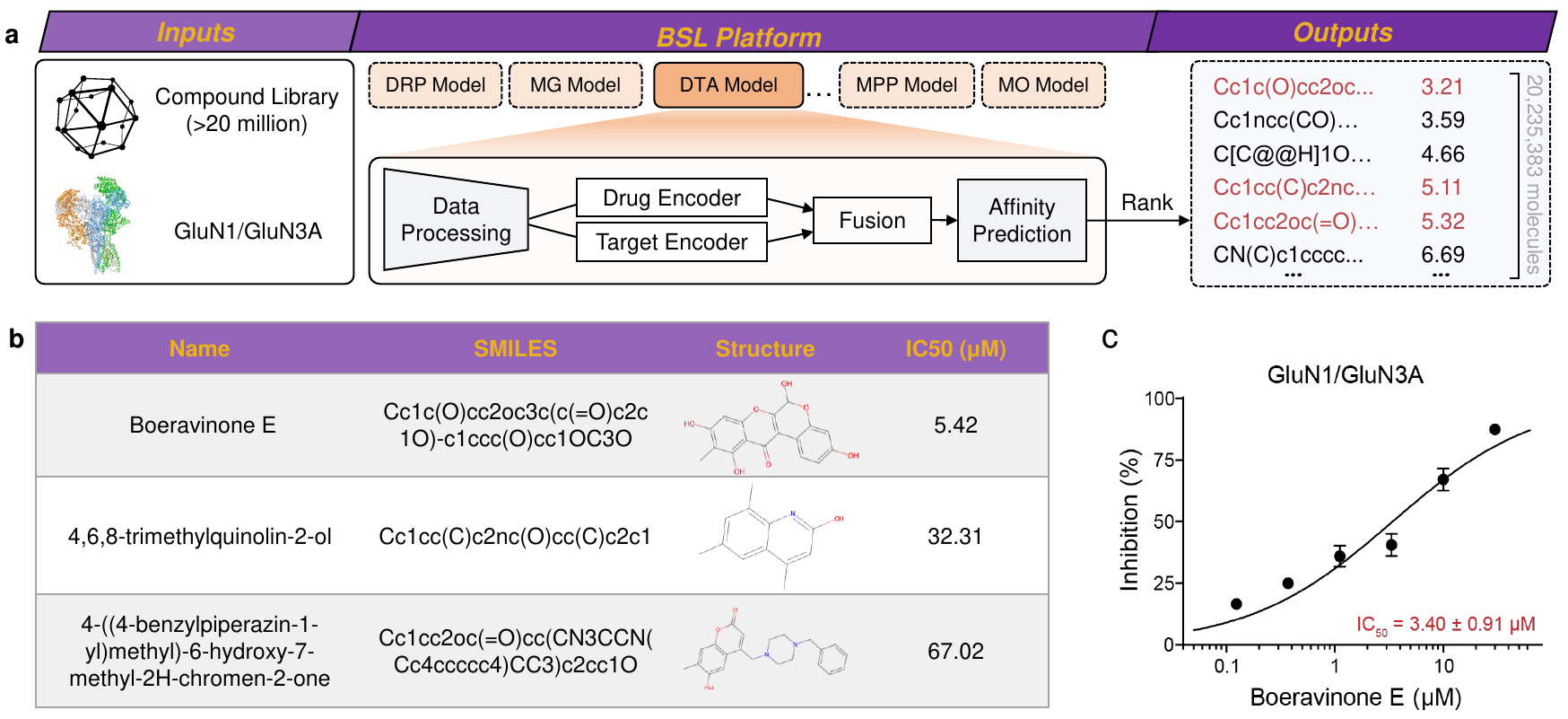}
    \caption{The drug screening process targeting GluN1/GluN3A in BSL was conducted using the AI-driven CLG-DTA model.}
    \label{fig:BSL_APS}
\end{figure*}

\subsection{Drug Cell Response Prediction Task}

The drug response prediction (DRP) task predicts how different compounds will perform in biological contexts, facilitating the prioritization of drug candidates. However, many existing methods struggle with limitations in extracting complex molecular structures and are limited in handling OOD samples, remaining unable to address generalization challenges. To tackle the OOD problems, BSL incorporates three innovative methods, all of which are specifically designed to address OOD issues, as illustrated in Fig.~\ref{fig:main}c. TransEDRP \cite{TransEDRP} utilizes a dual transformer architecture that integrates edge embeddings to effectively capture the pharmacochemical properties of drug molecules alongside genomic sequences from cell lines. MSDA \cite{MSDA} introduces a zero-shot learning paradigm tailored for preclinical drug screening. It designs a multi-source domain selector to select multiple drug domains that are similar to the target domain from the training dataset. The set of molecules $D^{S \rightarrow T}_i$ from source $S$ are selected to target $T$ as shown below:
\begin{equation}
    \begin{aligned}
        & \left\{
        \begin{aligned}
            & D^{S \rightarrow T}_i = \text{Top}_K(\textbf{W}{D_{ij}}) \\
            & \textbf{W}D_{ij} = \inf\mathbb{E}_{(x,y) \sim \gamma}[  ||x-y||], \\
        \end{aligned}
        \right.
    \end{aligned}
\end{equation}
where $\text{Top}_K(\cdot)$ denotes the operation of sorting from smallest to largest order and fetching the first $K$ elements. The Wasserstein distance $\textbf{W}D_{ij}$ is a distance function defined between probability distributions on a given metric space, where $\gamma$ denotes the joint distribution of the set $(x,y)$.
CLDR \cite{CLDR} employs a contrastive learning model with natural language supervision, transforming regression labels into text and integrating a common-sense numerical knowledge graph.

\subsection{Drug-Drug Interaction Prediction Task}

Drug-drug interaction prediction aims to identify potential interactions between two drugs. Conventional molecular representation approaches often fail to capture multi-granular knowledge, such as atomic-level details alongside pharmacological context \cite{wan2023molecules}, and struggle with handling the molecular heterogeneity present in complex drug interactions \cite{li2023dsn}. To address these challenges, our BSL incorporates the KCHML. As shown in Fig.~\ref{fig:main}d, it works by first creating comprehensive individual representations for each drug, combining their structural, chemical, and pharmacological information. Then, it uses a neural network to model the complex interactions between these drug pairs, effectively leveraging knowledge gained during pretraining to identify potential DDIs accurately. The graph embedding $\textbf{h}_G$ can be obtained as follows:
\begin{equation}
    \begin{aligned}
        & \left\{
        \begin{aligned}
            & \textbf{m}_{v_i}, \textbf{h}_{v_i} = \text{AGG}(\{ e_{pi} | v_p \}), \text{UPD}(\textbf{h}_{v_i}, \textbf{m}_{v_i}) \\
            & \textbf{m}_{e_{ij}}, \textbf{h}_{e_{ij}} = \text{AGG}(\{ v_i \} \cup \{ e_{pi}: v_p \}), \text{UPD}(\textbf{h}_{e_{ij}}, \textbf{m}_{e_{ij}}) \\
            & \textbf{h}_G = \text{READOUT}(\{ \textbf{h}_{v_i} \}, \{ \textbf{h}_{e_{ij}} \}). \\
        \end{aligned}
        \right.
    \end{aligned}
\end{equation}
where the $\textbf{h}_{v_i}$ and $\textbf{h}_{e_{ij}}$ respectively represent node and edge features, $\text{AGG}$ denotes aggregating neighborhood features, and $\text{UPD}$ denotes updating process.

\subsection{Drug Retrosynthesis Task}

Retrosynthesis prediction takes a product molecule as input and outputs the possible reactants for that product. Existing methods often fail to strictly follow molecular sequence rules during inference, which may result in structures that are not chemically accurate. To deeply integrate multimodal information and improve the accuracy and validity of predictions, we incorporate CFC-Retro \cite{zeng2025enhancing}, a template-free retrosynthesis prediction model, into the BSL. As shown in Fig.~\ref{fig:main}h, CFC-Retro integrates sequence and graph representations via a dual-branch encoder with cross-modal attention. The cross-modal fine-grained contrastive learning strategy aligns atomic-level representations of unchanged substructures across different modalities and pre-/post-reaction states, without requiring handcrafted alignments or data constraints.

\begin{figure*}[ht!]
    \centering
    \includegraphics[width=0.99\linewidth]{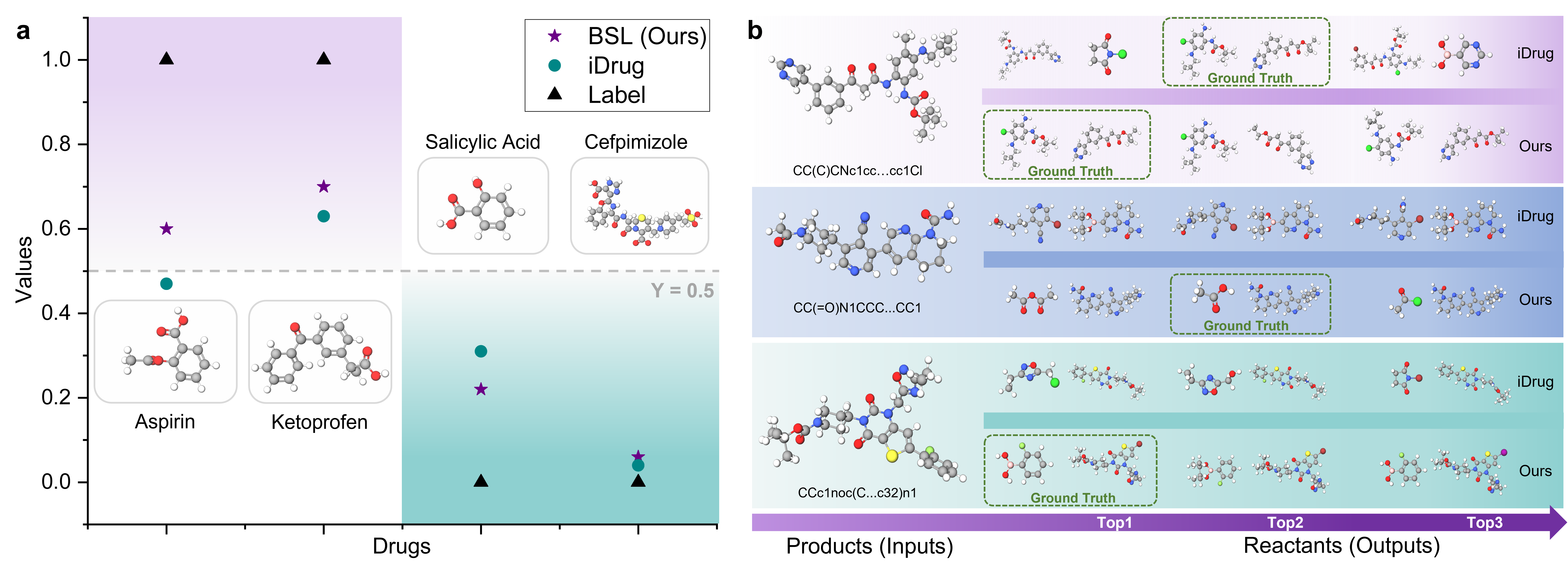}
    \caption{Case study results comparing the BBBP classification and retrosynthesis predictions of two platforms, BSL and iDrug, on OOD drug molecules.}
    \label{fig:case_study_retrosynthesis}
\end{figure*}

\section{Experiments}

\subsection{Comparative Analysis}
To evaluate BSL's capabilities, we conducted a comprehensive benchmarking study. \textbf{At the platform level}, BSL was compared with existing available platforms, as shown in Table \ref{tab:comparison}. \textbf{At the task level}, BSL was benchmarked against SOTA models across seven representative tasks, as shown in Table \ref{tab:comparison2}, using standardized datasets and widely adopted evaluation metrics.


For the MG task, we employed \textit{fréchet chemNet distance} (FCD) \cite{preuer2018frechet} and \textit{maximum mean discrepancy} (MMD) \cite{costa2010fast}. For the MO task, we used \textit{Success Rate} to measure the proportion of cell lines for which at least one improved molecule could be generated. For classification and regression tasks, we used standard metrics, such as the Pearson correlation coefficient (PCC) and mean absolute error (MAE), to ensure consistent and fair evaluation across tasks. All metrics follow standard definitions and calculation protocols as described in the original papers, and detailed information is provided in the Supplementary Materials.

As shown in Table \ref{tab:comparison2}, BSL consistently outperforms specialized baselines across multiple tasks, demonstrating both high predictive accuracy and rapid inference. These baselines represent SOTA models developed in the past three years for individual tasks. \textbf{ To the best of our knowledge, BSL represents the most publicly accessible end-to-end platform with the broadest task coverage and highest predictive accuracy to date.} Its release is expected to facilitate more efficient and accessible AI-powered solutions in pharmaceutical research.

\subsection{Application targeting GluN1/GluN3A}

The GluN1/GluN3A subtype of NMDA receptors, implicated in neurological disorders such as stroke and Alzheimer's disease. However, in the absence of such structural information for GluN1/GluN3A, conventional approaches become infeasible. To overcome this limitation, we leveraged the BSL platform to perform sequence-based virtual screening, bypassing the need for resolved crystal structures. The workflow is shown in Fig. \ref{fig:BSL_APS} a. The input includes the amino acid sequence of the GluN1/GluN3A receptor and a large-scale small-molecule compound library. These data are processed through the BSL platform, which utilizes CLG-DTA to predict compound–target binding affinities and prioritize candidates accordingly. The top-ranked molecules are then subjected to \textit{in vitro} validation using electrophysiological assays. Finally, we successfully identified three novel compounds exhibiting clear bioactivity against the GluN1/GluN3A receptor in \textit{in vitro} electrophysiological experiments (Fig.~\ref{fig:BSL_APS}b and c). These findings underscore the effectiveness of BSL in facilitating drug discovery in challenging settings where structural information is limited or unavailable.


\subsection{Case study}



To assess the OOD performance of our MPP model, we tested four drugs absent from public datasets \cite{Dehnbostel2024} using both the BSL and iDrug platforms for blood–brain barrier permeability (BBBP) prediction. The prediction results are shown in Fig. \ref{fig:case_study_retrosynthesis}a. BSL correctly predicted the classification of all drugs, while the iDrug platform achieved an accuracy of 75\%. Moreover, BSL's predicted probability scores were generally closer to the true labels, exhibiting higher confidence. These results validate the high accuracy and reliability of BSL in the molecular property prediction task.

To evaluate the practical performance of our retrosynthesis model, we conducted a case study on three structurally diverse target compounds (details in Supplementary Materials). As shown in Fig. \ref{fig:case_study_retrosynthesis}b, the model accurately identified the ground-truth reactants for all three cases, ranking them within the top-2 predictions, and twice at top-1. In contrast, the iDrug platform succeeded in only one case and failed in the other two. These results highlight the superior robustness of our model when applied to complex compounds.

\section{Conclusion}

This paper introduces the BSL, a comprehensive virtual screening tool focused on drug discovery. Our goal is to establish BSL as an essential resource for drug research, aiding scientists in faster, more accurate drug identification and development. The core functions of the BSL platform encompass all stages of drug development, including drug design, property prediction, and virtual screening. Equipped with high-precision predictive models, outstanding zero-shot generalization, and support for multimodal input, the platform effectively addresses a variety of challenges in drug development. Overall, BSL provides researchers with powerful, user-friendly tools that greatly increase the efficiency of drug discovery, driving progress in drug development. 



\clearpage

\bibliography{cas-refs}

\begin{thebibliography}{62}
\providecommand{\natexlab}[1]{#1}

\bibitem[{{\AA}qvist, Medina, and Samuelsson(1994)}]{cadd}
{\AA}qvist, J.; Medina, C.; and Samuelsson, J.-E. 1994.
\newblock A new method for predicting binding affinity in computer-aided drug design.
\newblock \emph{Protein Engineering, Design and Selection}, 7(3): 385--391.

\bibitem[{Ballester and Mitchell(2010)}]{ballester2010machine}
Ballester, P.~J.; and Mitchell, J.~B. 2010.
\newblock A machine learning approach to predicting protein--ligand binding affinity with applications to molecular docking.
\newblock \emph{Bioinformatics}, 26(9): 1169--1175.

\bibitem[{Catacutan et~al.(2024)Catacutan, Alexander, Arnold, and Stokes}]{catacutan2024machine}
Catacutan, D.~B.; Alexander, J.; Arnold, A.; and Stokes, J.~M. 2024.
\newblock Machine learning in preclinical drug discovery.
\newblock \emph{Nature Chemical Biology}, 20(8): 960--973.

\bibitem[{Chu et~al.(2023)Chu, Nguyen, Hai, Nguyen, and Nguyen}]{chu2023graphtsdrp}
Chu, T.; Nguyen, T.~T.; Hai, B.~D.; Nguyen, Q.~H.; and Nguyen, T. 2023.
\newblock Graph Transformer for Drug Response Prediction.
\newblock \emph{IEEE/ACM Transactions on Computational Biology and Bioinformatics}, 20(2): 1065--1072.

\bibitem[{Chu et~al.(2022)Chu, Huang, Fu, Quan, Zhou, Liu, and Zhang}]{chu2022hierarchical}
Chu, Z.; Huang, F.; Fu, H.; Quan, Y.; Zhou, X.; Liu, S.; and Zhang, W. 2022.
\newblock Hierarchical graph representation learning for the prediction of drug-target binding affinity.
\newblock \emph{Information Sciences}, 613: 507--523.

\bibitem[{Congreve, Dias et~al.(2014)}]{congreve2014structure}
Congreve, M.; Dias, J.~M.; et~al. 2014.
\newblock Structure-based drug design for G protein-coupled receptors.
\newblock \emph{Progress in medicinal chemistry}, 53: 1--63.

\bibitem[{Costa and De~Grave(2010)}]{costa2010fast}
Costa, F.; and De~Grave, K. 2010.
\newblock Fast neighborhood subgraph pairwise distance kernel.
\newblock In \emph{Proceedings of the 26th International Conference on Machine Learning}, 255--262. Omnipress; Madison, WI, USA.

\bibitem[{Dehnbostel et~al.(2024)Dehnbostel, Dixit, Preissner, and Banerjee}]{Dehnbostel2024}
Dehnbostel, F.~O.; Dixit, V.~A.; Preissner, R.; and Banerjee, P. 2024.
\newblock Non-animal models for blood--brain barrier permeability evaluation of drug-like compounds.
\newblock \emph{Scientific Reports}, 14(1): 8908.

\bibitem[{Fang et~al.(2022)Fang, Zhang, Yang, Zhuang, Deng, Zhang, Qin, Chen, Fan, and Chen}]{fang2022molecular}
Fang, Y.; Zhang, Q.; Yang, H.; Zhuang, X.; Deng, S.; Zhang, W.; Qin, M.; Chen, Z.; Fan, X.; and Chen, H. 2022.
\newblock Molecular contrastive learning with chemical element knowledge graph.
\newblock In \emph{Proceedings of the AAAI Conference on Artificial Intelligence}, volume~36, 3968--3976.

\bibitem[{Fine et~al.(2019)Fine, Lackner, Samudrala, and Chopra}]{fine2019computational}
Fine, J.; Lackner, R.; Samudrala, R.; and Chopra, G. 2019.
\newblock Computational chemoproteomics to understand the role of selected psychoactives in treating mental health indications.
\newblock \emph{Scientific reports}, 9(1): 13155.

\bibitem[{Gadioli et~al.(2022)Gadioli, Vitali, Ficarelli, Latini, Manelfi, Talarico, Silvano, Beccari, and Palermo}]{gadioli2022extreme}
Gadioli, D.; Vitali, E.; Ficarelli, F.; Latini, C.; Manelfi, C.; Talarico, C.; Silvano, C.; Beccari, A.~R.; and Palermo, G. 2022.
\newblock An extreme-scale virtual screening platform for drug discovery.
\newblock In \emph{Proceedings of the 19th ACM International Conference on Computing Frontiers}, 211--212.

\bibitem[{Guedes et~al.(2024)Guedes, da~Silva, Galheigo, Krempser, de~Magalh{\~a}es, Barbosa, and Dardenne}]{guedes2024dockthor}
Guedes, I.~A.; da~Silva, M. M.~P.; Galheigo, M.; Krempser, E.; de~Magalh{\~a}es, C.~S.; Barbosa, H. J.~C.; and Dardenne, L.~E. 2024.
\newblock DockThor-VS: A Free Platform for Receptor-Ligand Virtual Screening.
\newblock \emph{Journal of Molecular Biology}, 168548.

\bibitem[{Hu et~al.(2025)Hu, Li, Hu, Xiong, Cai, and Hu}]{huDDIRMCD}
Hu, C.; Li, K.; Hu, L.; Xiong, Y.; Cai, X.; and Hu, W. 2025.
\newblock Collaborative Drug Design Based on A Drug-Drug Interaction-Guided Diffusion Model.
\newblock In \emph{2025 28th International Conference on Computer Supported Cooperative Work in Design (CSCWD)}, 1374--1380.

\bibitem[{Huang et~al.(2023)Huang, Sun, Du, and Lv}]{huang2023conditional}
Huang, H.; Sun, L.; Du, B.; and Lv, W. 2023.
\newblock Conditional diffusion based on discrete graph structures for molecular graph generation.
\newblock In \emph{Proceedings of the AAAI Conference on Artificial Intelligence}, volume~37, 4302--4311.

\bibitem[{Inukai et~al.(2024)Inukai, Yamato, Akiyama, and Sakakibara}]{inukai2024tree}
Inukai, T.; Yamato, A.; Akiyama, M.; and Sakakibara, Y. 2024.
\newblock A Tree-Transformer based VAE with fragment tokenization for large chemical models.
\newblock \emph{ChemRxiv preprint}.

\bibitem[{Ivanenkov et~al.(2023)Ivanenkov, Polykovskiy, Bezrukov et~al.}]{ivanenkov2023chemistry42}
Ivanenkov, Y.~A.; Polykovskiy, D.; Bezrukov, D.; et~al. 2023.
\newblock Chemistry42: an AI-driven platform for molecular design and optimization.
\newblock \emph{Journal of Chemical Information and Modeling}, 63(3): 695--701.

\bibitem[{Jin et~al.(2023)Jin, Wang, Shi, Bao, Wang, Zhang, Pan, Li, Yao, Liu et~al.}]{jin2023fflom}
Jin, J.; Wang, D.; Shi, G.; Bao, J.; Wang, J.; Zhang, H.; Pan, P.; Li, D.; Yao, X.; Liu, H.; et~al. 2023.
\newblock FFLOM: a flow-based autoregressive model for fragment-to-lead optimization.
\newblock \emph{Journal of Medicinal Chemistry}, 66(15): 10808--10823.

\bibitem[{Jo, Kim, and Hwang(2023)}]{jo2023graph}
Jo, J.; Kim, D.; and Hwang, S.~J. 2023.
\newblock Graph generation with destination-driven diffusion mixture.
\newblock In \emph{ICLR 2023-Machine Learning for Drug Discovery workshop}.

\bibitem[{Keda et~al.(2024)Keda, Zewen, Zhen et~al.}]{keda2024molprophet}
Keda, Y.; Zewen, X.; Zhen, L.; et~al. 2024.
\newblock MolProphet: A One-Stop, General Purpose, and AI-Based Platform for the Early Stages of Drug Discovery.
\newblock \emph{Journal of Chemical Information and Modeling}, 64(8): 2941--2947.

\bibitem[{Kuhn, Braslavsky, and Schmidt(2004)}]{iupac}
Kuhn, H.; Braslavsky, S.; and Schmidt, R. 2004.
\newblock Chemical actinometry (IUPAC technical report).
\newblock \emph{Pure and Applied Chemistry}, 76(12): 2105--2146.

\bibitem[{Kun and Wenbin(2022)}]{TransEDRP}
Kun, L.; and Wenbin, H. 2022.
\newblock TransEDRP: Dual Transformer model with Edge Emdedded for Drug Respond Prediction.
\newblock arXiv:2210.17401.

\bibitem[{Lavecchia et~al.(2024)}]{lavecchia2024advancing}
Lavecchia, A.; et~al. 2024.
\newblock Advancing drug discovery with deep attention neural networks.
\newblock \emph{DRUG DISCOVERY TODAY}, 29(8).

\bibitem[{Lee, Jo, and Hwang(2023)}]{lee2023exploring}
Lee, S.; Jo, J.; and Hwang, S.~J. 2023.
\newblock Exploring chemical space with score-based out-of-distribution generation.
\newblock In \emph{International Conference on Machine Learning}, 18872--18892. PMLR.

\bibitem[{Li et~al.(2024{\natexlab{a}})Li, Cai, Wu, Du, and Hu}]{li2024fragment}
Li, K.; Cai, X.; Wu, J.; Du, B.; and Hu, W. 2024{\natexlab{a}}.
\newblock Fragment-Masked Molecular Optimization.
\newblock \emph{arXiv preprint arXiv:2408.09106}.

\bibitem[{Li et~al.(2024{\natexlab{b}})Li, Gong, Wu, and Hu}]{CLDR}
Li, K.; Gong, X.; Wu, J.; and Hu, W. 2024{\natexlab{b}}.
\newblock Contrastive Learning Drug Response Models from Natural Language Supervision.
\newblock In Larson, K., ed., \emph{Proceedings of the Thirty-Third International Joint Conference on Artificial Intelligence, {IJCAI-24}}, 2126--2134. International Joint Conferences on Artificial Intelligence Organization.
\newblock Main Track.

\bibitem[{Li et~al.(2025{\natexlab{a}})Li, Hu, Cai, Wu, and Hu}]{mevon}
Li, K.; Hu, L.; Cai, X.; Wu, J.; and Hu, W. 2025{\natexlab{a}}.
\newblock Can Molecular Evolution Mechanism Enhance Molecular Representation?
\newblock \emph{arXiv preprint arXiv:2501.15799}.

\bibitem[{Li et~al.(2024{\natexlab{c}})Li, Liu, Luo, Cai, Wu, and Hu}]{MSDA}
Li, K.; Liu, W.; Luo, Y.; Cai, X.; Wu, J.; and Hu, W. 2024{\natexlab{c}}.
\newblock Zero-shot Learning for Preclinical Drug Screening.
\newblock In Larson, K., ed., \emph{Proceedings of the Thirty-Third International Joint Conference on Artificial Intelligence, {IJCAI-24}}, 2117--2125. International Joint Conferences on Artificial Intelligence Organization.
\newblock Main Track.

\bibitem[{Li et~al.(2022)Li, Wu, Du, Petoukhov, Xu, Xiao, and Hu}]{li2022transedrp}
Li, K.; Wu, J.; Du, B.; Petoukhov, S.~V.; Xu, H.; Xiao, Z.; and Hu, W. 2022.
\newblock TransEDRP: Dual Transformer Model with Edge Embedded for Drug Respond Prediction.
\newblock \emph{arXiv preprint arXiv:2210.17401}.

\bibitem[{Li et~al.(2025{\natexlab{b}})Li, Zeng, Xiong, Wu, Fang, Qu, Zhu, Du, Gao, and Hu}]{li2025contrastive}
Li, K.; Zeng, Y.; Xiong, Y.-d.; Wu, H.-c.; Fang, S.; Qu, Z.-y.; Zhu, Y.; Du, B.; Gao, Z.-b.; and Hu, W.-b. 2025{\natexlab{b}}.
\newblock Contrastive learning-based drug screening model for GluN1/GluN3A inhibitors.
\newblock \emph{Acta Pharmacologica Sinica}, 1--13.

\bibitem[{Li et~al.(2023)Li, Zhu, Shao, Zeng, Wang, and Liu}]{li2023dsn}
Li, Z.; Zhu, S.; Shao, B.; Zeng, X.; Wang, T.; and Liu, T.-Y. 2023.
\newblock DSN-DDI: an accurate and generalized framework for drug--drug interaction prediction by dual-view representation learning.
\newblock \emph{Briefings in Bioinformatics}, 24(1): bbac597.

\bibitem[{Liao and Smidt(2023)}]{equiformer}
Liao, Y.-L.; and Smidt, T. 2023.
\newblock Equiformer: Equivariant Graph Attention Transformer for 3D Atomistic Graphs.
\newblock In \emph{The Eleventh International Conference on Learning Representations}.

\bibitem[{Lin et~al.(2023)Lin, Xu, Xiong, Zhang, Ni, Ni, Chang, Pan, Wang, Yu et~al.}]{lin2022pangu}
Lin, X.; Xu, C.; Xiong, Z.; Zhang, X.; Ni, N.; Ni, B.; Chang, J.; Pan, R.; Wang, Z.; Yu, F.; et~al. 2023.
\newblock PanGu Drug Model: learn a molecule like a human.
\newblock \emph{Science China. Life sciences}, 66(4): 879--882.

\bibitem[{Liu et~al.(2024)Liu, Wu, Wang, Deng, and Zhou}]{liu2024higraphdti}
Liu, B.; Wu, S.; Wang, J.; Deng, X.; and Zhou, A. 2024.
\newblock Higraphdti: Hierarchical graph representation learning for drug-target interaction prediction.
\newblock In \emph{Joint European Conference on Machine Learning and Knowledge Discovery in Databases}, 354--370. Springer.

\bibitem[{Ma and Lei(2023)}]{ma2023dual}
Ma, M.; and Lei, X. 2023.
\newblock A dual graph neural network for drug--drug interactions prediction based on molecular structure and interactions.
\newblock \emph{PLOS Computational Biology}, 19(1): e1010812.

\bibitem[{Mao et~al.(2021)Mao, Xiao, Xu, Rong, Huang, and Zhao}]{mao2021molecular}
Mao, K.; Xiao, X.; Xu, T.; Rong, Y.; Huang, J.; and Zhao, P. 2021.
\newblock Molecular graph enhanced transformer for retrosynthesis prediction.
\newblock \emph{Neurocomputing}, 457: 193--202.

\bibitem[{Mishra et~al.(2022)Mishra, Malik, Hachiya, J{\"u}rgenson, Namba, Posner, Kamanu, Koido, Le~Grand, Shi et~al.}]{mishra2022stroke}
Mishra, A.; Malik, R.; Hachiya, T.; J{\"u}rgenson, T.; Namba, S.; Posner, D.~C.; Kamanu, F.~K.; Koido, M.; Le~Grand, Q.; Shi, M.; et~al. 2022.
\newblock Stroke genetics informs drug discovery and risk prediction across ancestries.
\newblock \emph{Nature}, 611(7934): 115--123.

\bibitem[{Mock et~al.(2023)Mock, Edavettal, Langmead, and Russell}]{mock2023ai}
Mock, M.; Edavettal, S.; Langmead, C.; and Russell, A. 2023.
\newblock AI can help to speed up drug discovery—but only if we give it the right data.
\newblock \emph{Nature}, 621(7979): 467--470.

\bibitem[{Morehead and Cheng(2024)}]{MG4}
Morehead, A.; and Cheng, J. 2024.
\newblock Geometry-complete diffusion for 3D molecule generation and optimization.
\newblock \emph{Communications Chemistry}, 7(1): 150.

\bibitem[{Nyamabo et~al.(2022)Nyamabo, Yu, Liu, and Shi}]{nyamabo2022drug}
Nyamabo, A.~K.; Yu, H.; Liu, Z.; and Shi, J.-Y. 2022.
\newblock Drug--drug interaction prediction with learnable size-adaptive molecular substructures.
\newblock \emph{Briefings in Bioinformatics}, 23(1): bbab441.

\bibitem[{Peng et~al.(2024)Peng, Liu, Chen, Liao, Mao, and Zhou}]{peng2024mgndti}
Peng, L.; Liu, X.; Chen, M.; Liao, W.; Mao, J.; and Zhou, L. 2024.
\newblock MGNDTI: a drug-target interaction prediction framework based on multimodal representation learning and the gating mechanism.
\newblock \emph{Journal of Chemical Information and Modeling}, 64(16): 6684--6698.

\bibitem[{Preuer et~al.(2018)Preuer, Renz, Unterthiner, Hochreiter, and Klambauer}]{preuer2018frechet}
Preuer, K.; Renz, P.; Unterthiner, T.; Hochreiter, S.; and Klambauer, G. 2018.
\newblock Fr{\'e}chet ChemNet distance: a metric for generative models for molecules in drug discovery.
\newblock \emph{Journal of chemical information and modeling}, 58(9): 1736--1741.

\bibitem[{Rong et~al.(2020)Rong, Bian, Xu, Xie, Wei, Huang, and Huang}]{rong2020self}
Rong, Y.; Bian, Y.; Xu, T.; Xie, W.; Wei, Y.; Huang, W.; and Huang, J. 2020.
\newblock Self-supervised graph transformer on large-scale molecular data.
\newblock \emph{Advances in neural information processing systems}, 33: 12559--12571.

\bibitem[{Rusinko et~al.(2023)Rusinko, Rezaei, Friedrich et~al.}]{rusinko2023aiddison}
Rusinko, A.; Rezaei, M.; Friedrich, L.; et~al. 2023.
\newblock AIDDISON: Empowering Drug Discovery with AI/ML and CADD Tools in a Secure, Web-Based SaaS Platform.
\newblock \emph{Journal of Chemical Information and Modeling}, 64(1): 3--8.

\bibitem[{Shen et~al.(2024)Shen, Song, Hsieh et~al.}]{shen2024drugflow}
Shen, C.; Song, J.; Hsieh, C.~Y.; et~al. 2024.
\newblock DrugFlow: An AI-Driven One-Stop Platform for Innovative Drug Discovery.
\newblock \emph{Journal of Chemical Information and Modeling}, 64(14): 5381--5391.

\bibitem[{Shin et~al.(2024)Shin, Son, Han, Kam et~al.}]{shin2024dymol}
Shin, D.-H.; Son, Y.-H.; Han, J.-W.; Kam, T.-E.; et~al. 2024.
\newblock Dymol: Dynamic many-objective molecular optimization with objective decomposition and progressive optimization.
\newblock In \emph{ICLR 2024 Workshop on Generative and Experimental Perspectives for Biomolecular Design}.

\bibitem[{Sun et~al.(2022)Sun, Xing, Meng, Wang, Chen, and Zhou}]{sun2022molsearch}
Sun, M.; Xing, J.; Meng, H.; Wang, H.; Chen, B.; and Zhou, J. 2022.
\newblock Molsearch: search-based multi-objective molecular generation and property optimization.
\newblock In \emph{Proceedings of the 28th ACM SIGKDD conference on knowledge discovery and data mining}, 4724--4732.

\bibitem[{Tang et~al.(2022)Tang, Jensen, Houang, McRobb, Bhat, Svensson, Bochevarov, Day, Dahlgren, Bell et~al.}]{tang2022discoveryXDE}
Tang, H.; Jensen, K.; Houang, E.; McRobb, F.~M.; Bhat, S.; Svensson, M.; Bochevarov, A.; Day, T.; Dahlgren, M.~K.; Bell, J.~A.; et~al. 2022.
\newblock Discovery of a Novel Class of D-Amino Acid Oxidase Inhibitors Using the Schr\"odinger Computational Platform.
\newblock \emph{Journal of Medicinal Chemistry}, 65(9): 6775--6802.

\bibitem[{Volokhova et~al.(2024)Volokhova, Koziarski, Hern{\'a}ndez-Garc{\'\i}a, Liu, Miret, Lemos, Thiede, Yan, Aspuru-Guzik, and Bengio}]{MG3}
Volokhova, A.; Koziarski, M.; Hern{\'a}ndez-Garc{\'\i}a, A.; Liu, C.-H.; Miret, S.; Lemos, P.; Thiede, L.; Yan, Z.; Aspuru-Guzik, A.; and Bengio, Y. 2024.
\newblock Towards equilibrium molecular conformation generation with GFlowNets.
\newblock \emph{Digital Discovery}, 3(5): 1038--1047.

\bibitem[{Wan et~al.(2023)Wan, Wu, Hou, Hsieh, and Jia}]{wan2023molecules}
Wan, Y.; Wu, J.; Hou, T.; Hsieh, C.-Y.; and Jia, X. 2023.
\newblock From molecules to scaffolds to functional groups: building context-dependent molecular representation via multi-channel learning.
\newblock \emph{arXiv preprint arXiv:2311.02798}.

\bibitem[{Wang et~al.(2022)Wang, Liu, Lin, Liu, and Ji}]{NEURIPS2022_0418973e}
Wang, L.; Liu, Y.; Lin, Y.; Liu, H.; and Ji, S. 2022.
\newblock ComENet: Towards Complete and Efficient Message Passing for 3D Molecular Graphs.
\newblock In Koyejo, S.; Mohamed, S.; Agarwal, A.; Belgrave, D.; Cho, K.; and Oh, A., eds., \emph{Advances in Neural Information Processing Systems}, volume~35, 650--664. Curran Associates, Inc.

\bibitem[{Wang et~al.(2023)Wang, Song, Zhang, Zhang, Liu, Ren, and Pang}]{wang2023msgnn}
Wang, S.; Song, X.; Zhang, Y.; Zhang, K.; Liu, Y.; Ren, C.; and Pang, S. 2023.
\newblock MSGNN-DTA: multi-scale topological feature fusion based on graph neural networks for drug--target binding affinity prediction.
\newblock \emph{International Journal of Molecular Sciences}, 24(9): 8326.

\bibitem[{Wang et~al.(2024)Wang, Wang, Li, He, Li, Wang, Zheng, Shao, and Liu}]{visnet}
Wang, Y.; Wang, T.; Li, S.; He, X.; Li, M.; Wang, Z.; Zheng, N.; Shao, B.; and Liu, T.-Y. 2024.
\newblock Enhancing geometric representations for molecules with equivariant vector-scalar interactive message passing.
\newblock \emph{Nature Communications}, 15(1): 313.

\bibitem[{Wieder et~al.(2020)Wieder, Kohlbacher, Kuenemann, Garon, Ducrot, Seidel, and Langer}]{wieder2020compact}
Wieder, O.; Kohlbacher, S.; Kuenemann, M.; Garon, A.; Ducrot, P.; Seidel, T.; and Langer, T. 2020.
\newblock A compact review of molecular property prediction with graph neural networks.
\newblock \emph{Drug Discovery Today: Technologies}, 37: 1--12.

\bibitem[{Wu et~al.(2022)Wu, Jin, Jiao, Wang, Li, and Pan}]{wu2022carbonai}
Wu, J.; Jin, K.; Jiao, Y.; Wang, X.; Li, S.; and Pan, L. 2022.
\newblock CarbonAI, A Non-Docking Deep learning based small molecule virtual screening platform.
\newblock \emph{ChemRxiv preprint}.

\bibitem[{Wu et~al.(2024)Wu, Zhang, Wang, Fu, Zhao, Wang, Du, Jiang, Deng, Cao et~al.}]{wu2024leveraging}
Wu, Z.; Zhang, O.; Wang, X.; Fu, L.; Zhao, H.; Wang, J.; Du, H.; Jiang, D.; Deng, Y.; Cao, D.; et~al. 2024.
\newblock Leveraging language model for advanced multiproperty molecular optimization via prompt engineering.
\newblock \emph{Nature Machine Intelligence}, 6(11): 1359--1369.

\bibitem[{Xiong et~al.(2024)Xiong, Li, Liu, Wu, Du, Pan, and Hu}]{xiong2024text}
Xiong, Y.; Li, K.; Liu, W.; Wu, J.; Du, B.; Pan, S.; and Hu, W. 2024.
\newblock Text-Guided Multi-Property Molecular Optimization with a Diffusion Language Model.
\newblock \emph{arXiv preprint arXiv:2410.13597}.

\bibitem[{Yan et~al.(2020)Yan, Ding, Zhao, Zheng, Yang, Yu, and Huang}]{yan2020retroxpert}
Yan, C.; Ding, Q.; Zhao, P.; Zheng, S.; Yang, J.; Yu, Y.; and Huang, J. 2020.
\newblock Retroxpert: Decompose retrosynthesis prediction like a chemist.
\newblock \emph{Advances in Neural Information Processing Systems}, 33: 11248--11258.

\bibitem[{Yin et~al.(2024)Yin, Chen, Hao, Pandiyan, Shao, and Wang}]{yin2024fotf}
Yin, Z.; Chen, Y.; Hao, Y.; Pandiyan, S.; Shao, J.; and Wang, L. 2024.
\newblock FOTF-CPI: A compound-protein interaction prediction transformer based on the fusion of optimal transport fragments.
\newblock \emph{Iscience}, 27(1).

\bibitem[{Zeng et~al.(2025)Zeng, Yao, Song, Wu, and Hu}]{zeng2025enhancing}
Zeng, J.; Yao, Z.; Song, P.; Wu, J.; and Hu, W. 2025.
\newblock Enhancing Template-Free Retrosynthesis Prediction with Cross-Modal Fine-Grained Contrastive Learning.
\newblock In \emph{2025 International Joint Conference on Neural Networks (IJCNN)}. IEEE.

\bibitem[{Zheng et~al.(2019)Zheng, Rao, Zhang, Xu, and Yang}]{zheng2019predicting}
Zheng, S.; Rao, J.; Zhang, Z.; Xu, J.; and Yang, Y. 2019.
\newblock Predicting retrosynthetic reactions using self-corrected transformer neural networks.
\newblock \emph{Journal of chemical information and modeling}, 60(1): 47--55.

\bibitem[{Zhou et~al.(2024)Zhou, Rusnac, Park, Canzani, Nguyen, Stewart, Bush, Nguyen, Wulff, Yarov-Yarovoy et~al.}]{zhou2024artificial}
Zhou, G.; Rusnac, D.-V.; Park, H.; Canzani, D.; Nguyen, H.~M.; Stewart, L.; Bush, M.~F.; Nguyen, P.~T.; Wulff, H.; Yarov-Yarovoy, V.; et~al. 2024.
\newblock An artificial intelligence accelerated virtual screening platform for drug discovery.
\newblock \emph{Nature Communications}, 15(1): 7761.

\bibitem[{Zhu et~al.(2023)Zhu, Wu, Hu, Yan, Hou, Wu et~al.}]{zhu2023sample}
Zhu, Y.; Wu, J.; Hu, C.; Yan, J.; Hou, T.; Wu, J.; et~al. 2023.
\newblock Sample-efficient multi-objective molecular optimization with gflownets.
\newblock \emph{Advances in Neural Information Processing Systems}, 36: 79667--79684.

\end{thebibliography}

\clearpage
\makeatletter
\@ifundefined{isChecklistMainFile}{
  \newif\ifreproStandalone
  \reproStandalonetrue
}{
  \newif\ifreproStandalone
  \reproStandalonefalse
}
\makeatother

\ifreproStandalone
\documentclass[letterpaper]{article}
\usepackage[submission]{aaai2026}
\setlength{\pdfpagewidth}{8.5in}
\setlength{\pdfpageheight}{11in}
\usepackage{times}
\usepackage{helvet}
\usepackage{courier}
\usepackage{xcolor}
\frenchspacing

\begin{document}
\fi
\setlength{\leftmargini}{20pt}
\makeatletter\def\@listi{\leftmargin\leftmargini \topsep .5em \parsep .5em \itemsep .5em}
\def\@listii{\leftmargin\leftmarginii \labelwidth\leftmarginii \advance\labelwidth-\labelsep \topsep .4em \parsep .4em \itemsep .4em}
\def\@listiii{\leftmargin\leftmarginiii \labelwidth\leftmarginiii \advance\labelwidth-\labelsep \topsep .4em \parsep .4em \itemsep .4em}\makeatother

\setcounter{secnumdepth}{0}
\renewcommand\thesubsection{\arabic{subsection}}
\renewcommand\labelenumi{\thesubsection.\arabic{enumi}}

\newcounter{checksubsection}
\newcounter{checkitem}[checksubsection]

\newcommand{\checksubsection}[1]{%
  \refstepcounter{checksubsection}%
  \paragraph{\arabic{checksubsection}. #1}%
  \setcounter{checkitem}{0}%
}

\newcommand{\checkitem}{%
  \refstepcounter{checkitem}%
  \item[\arabic{checksubsection}.\arabic{checkitem}.]%
}
\newcommand{\question}[2]{\normalcolor\checkitem #1 #2 \color{blue}}
\newcommand{\ifyespoints}[1]{\makebox[0pt][l]{\hspace{-15pt}\normalcolor #1}}

\section*{Reproducibility Checklist}









\checksubsection{General Paper Structure}
\begin{itemize}

\question{Includes a conceptual outline and/or pseudocode description of AI methods introduced}{(yes/partial/no/NA)}
\textbf{NA}

\question{Clearly delineates statements that are opinions, hypothesis, and speculation from objective facts and results}{(yes/no)}
\textbf{yes}

\question{Provides well-marked pedagogical references for less-familiar readers to gain background necessary to replicate the paper}{(yes/no)}
\textbf{yes}

\end{itemize}
\checksubsection{Theoretical Contributions}
\begin{itemize}

\question{Does this paper make theoretical contributions?}{(yes/no)}
\textbf{no}

	\ifyespoints{\vspace{1.2em}If yes, please address the following points:}
        \begin{itemize}
	
	\question{All assumptions and restrictions are stated clearly and formally}{(yes/partial/no)}
	Type your response here

	\question{All novel claims are stated formally (e.g., in theorem statements)}{(yes/partial/no)}
	Type your response here

	\question{Proofs of all novel claims are included}{(yes/partial/no)}
	Type your response here

	\question{Proof sketches or intuitions are given for complex and/or novel results}{(yes/partial/no)}
	Type your response here

	\question{Appropriate citations to theoretical tools used are given}{(yes/partial/no)}
	Type your response here

	\question{All theoretical claims are demonstrated empirically to hold}{(yes/partial/no/NA)}
	Type your response here

	\question{All experimental code used to eliminate or disprove claims is included}{(yes/no/NA)}
	Type your response here
	
	\end{itemize}
\end{itemize}

\checksubsection{Dataset Usage}
\begin{itemize}

\question{Does this paper rely on one or more datasets?}{(yes/no)}
\textbf{yes}

\ifyespoints{If yes, please address the following points:}
\begin{itemize}

	\question{A motivation is given for why the experiments are conducted on the selected datasets}{(yes/partial/no/NA)}
	\textbf{yes}

	\question{All novel datasets introduced in this paper are included in a data appendix}{(yes/partial/no/NA)}
	\textbf{NA}

	\question{All novel datasets introduced in this paper will be made publicly available upon publication of the paper with a license that allows free usage for research purposes}{(yes/partial/no/NA)}
	\textbf{NA}

	\question{All datasets drawn from the existing literature (potentially including authors' own previously published work) are accompanied by appropriate citations}{(yes/no/NA)}
	\textbf{yes}

	\question{All datasets drawn from the existing literature (potentially including authors' own previously published work) are publicly available}{(yes/partial/no/NA)}
	\textbf{yes}

	\question{All datasets that are not publicly available are described in detail, with explanation why publicly available alternatives are not scientifically satisficing}{(yes/partial/no/NA)}
	\textbf{NA}

\end{itemize}
\end{itemize}

\checksubsection{Computational Experiments}
\begin{itemize}

\question{Does this paper include computational experiments?}{(yes/no)}
\textbf{yes}

\ifyespoints{If yes, please address the following points:}
\begin{itemize}

	\question{This paper states the number and range of values tried per (hyper-) parameter during development of the paper, along with the criterion used for selecting the final parameter setting}{(yes/partial/no/NA)}
	\textbf{partial}

	\question{Any code required for pre-processing data is included in the appendix}{(yes/partial/no)}
	\textbf{yes}

	\question{All source code required for conducting and analyzing the experiments is included in a code appendix}{(yes/partial/no)}
	\textbf{yes}

	\question{All source code required for conducting and analyzing the experiments will be made publicly available upon publication of the paper with a license that allows free usage for research purposes}{(yes/partial/no)}
	\textbf{yes}
        
	\question{All source code implementing new methods have comments detailing the implementation, with references to the paper where each step comes from}{(yes/partial/no)}
	\textbf{yes}

	\question{If an algorithm depends on randomness, then the method used for setting seeds is described in a way sufficient to allow replication of results}{(yes/partial/no/NA)}
	\textbf{yes}

	\question{This paper specifies the computing infrastructure used for running experiments (hardware and software), including GPU/CPU models; amount of memory; operating system; names and versions of relevant software libraries and frameworks}{(yes/partial/no)}
	\textbf{yes}

	\question{This paper formally describes evaluation metrics used and explains the motivation for choosing these metrics}{(yes/partial/no)}
	\textbf{yes}

	\question{This paper states the number of algorithm runs used to compute each reported result}{(yes/no)}
	\textbf{yes}

	\question{Analysis of experiments goes beyond single-dimensional summaries of performance (e.g., average; median) to include measures of variation, confidence, or other distributional information}{(yes/no)}
	\textbf{yes}

	\question{The significance of any improvement or decrease in performance is judged using appropriate statistical tests (e.g., Wilcoxon signed-rank)}{(yes/partial/no)}
	\textbf{yes}

	\question{This paper lists all final (hyper-)parameters used for each model/algorithm in the paper’s experiments}{(yes/partial/no/NA)}
	\textbf{yes}

\end{itemize}
\end{itemize}
\ifreproStandalone
\end{document}
\fi
\end{document}